%% file: main.tex
\documentclass[10pt,twocolumn,letterpaper]{article}

\usepackage[pagenumbers]{cvpr} 
\input{preamble}

\definecolor{cvprblue}{rgb}{0.21,0.49,0.74}
\usepackage[pagebackref,breaklinks,colorlinks,citecolor=cvprblue]{hyperref}
\usepackage{pifont}
\usepackage{color}
\usepackage{xcolor}
\usepackage{nicefrac}       %
\usepackage{multirow}
\usepackage{multicol}
\usepackage{dsfont}

\usepackage{colortbl}
\usepackage[percent]{overpic}

\def\paperID{*****} 
\def\confName{CVPR}
\def\confYear{2024}

\title{Projecting Gaussian Ellipsoids While Avoiding Affine Projection Approximation}

\author{
	Han Qi$^{1}$ \quad Tao Cai$^{2}$ \quad Xiyue Han$^{3}$ \vspace{8pt}\\
	$^{1}$Beijing Institute of Technology
}

\begin{document}

\twocolumn[{%
	\renewcommand\twocolumn[1][]{#1}%
	\maketitle
	\includegraphics[width=1.\linewidth]{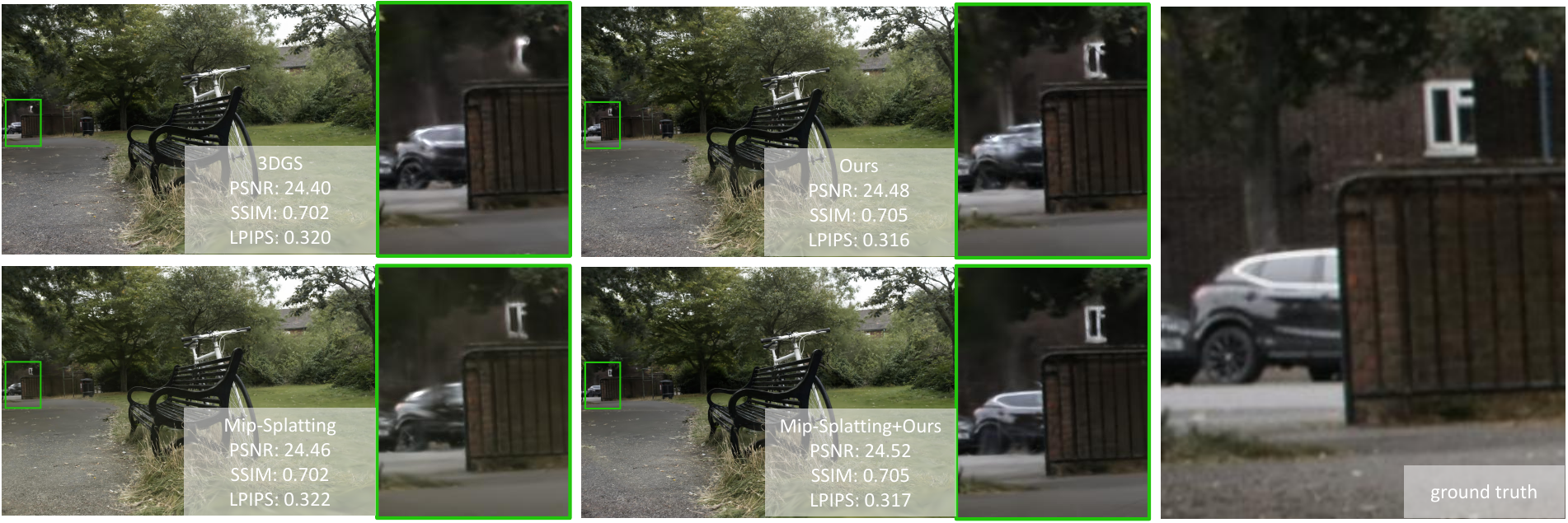}
	
    \captionof{figure}{Our method achieves a comprehensive improvement in rendering quality and rendering speed compared to 3D Gaussian splatting (3DGS)~\cite{3DGS}. We propose an ellipsoid-based projection method to replace the Jacobian of the affine approximation of the projection transformation in 3DGS. Our ellipsoid-based projection method can be applied to any 3DGS-based work to enhance rendering quality. This figure shows the rendering results of applying our method to 3DGS and Mip-Splatting~\cite{anti-aliasing1} in the scene \textit{bicycle} of Mip-NeRF360 dataset~\cite{render_quality2}, in which the rendering quality is enhanced with less blur and artifacts.}
	\label{fig:teaser}
	\vspace{2em}
}]

\input{sec/0.abstract} 
\input{sec/1.Introduction}
\input{sec/2.RelatedWorks}
\input{sec/3.Preliminaries}
\input{sec/4.ProposedMethod}

\input{sec/5.Experiments}
\input{sec/6.Conclusion}


{
    \small
    \bibliographystyle{ieeenat_fullname}
    \bibliography{main}
}


\end{document}

%% file: preamble.tex
\usepackage[dvipsnames]{xcolor}

\makeatletter
\DeclareRobustCommand\onedot{\futurelet\@let@token\@onedot}
\def\@onedot{\ifx\@let@token.\else.\null\fi\xspace}

\makeatother

\definecolor{yellow}{rgb}{1, 1, 0.7}
\definecolor{orange}{rgb}{1, 0.85, 0.7}
\definecolor{tablered}{rgb}{1, 0.7, 0.7}
\definecolor{red}{rgb}{1, 0, 0}

\definecolor{wincolor}{rgb}{0.85, 0.0, 0.0}

\definecolor{darkyellow}{rgb}{0.8, 0.8, 0.5}
\definecolor{darkred}{rgb}{0.7, 0.3, 0.3}
\definecolor{darkgreen}{rgb}{0.3, 0.7, 0.3}
\definecolor{blue}{rgb}{0.251, 0.498, 0.824}
\definecolor{green}{rgb}{0, 1.0, 0}
\definecolor{pink}{rgb}{1, 0.4, 0.7}

\definecolor{realred}{rgb}{0.95, 0.1, 0.0}

%% file: sec/0.abstract.tex
\begin{abstract}
Recently, 3D Gaussian Splatting has dominated novel-view synthesis with its real-time rendering speed and state-of-the-art rendering quality. However, during the rendering process, the use of the Jacobian of the affine approximation of the projection transformation leads to inevitable errors, resulting in blurriness, artifacts and a lack of scene consistency in the final rendered images. To address this issue, we introduce an ellipsoid-based projection method to calculate the projection of Gaussian ellipsoid onto the image plane, which is the primitive of 3D Gaussian Splatting. As our proposed ellipsoid-based projection method cannot handle Gaussian ellipsoids with camera origins inside them or parts lying below $z=0$ plane in the camera space, we designed a pre-filtering strategy. Experiments over multiple widely adopted benchmark datasets show that our ellipsoid-based projection method can enhance the rendering quality of 3D Gaussian Splatting and its extensions.
\end{abstract}

%% file: sec/1.Introduction.tex
\section{Introduction}

Novel View Synthesis (NVS) plays a crucial role in computer vision and computer graphics, with numerous applications, including robotics, virtual reality, and 3D gaming. One of the most influential works in this field is the Neural Radiance Field (NeRF)~\cite{NeRF}, proposed by Mildenhall et al. in 2020. NeRF utilizes a Multilayer Perceptron (MLP) to store geometric and appearance information of a scene and employs differentiable volume rendering \cite{volume_rendering1,volume_rendering2,volume_rendering3,volume_rendering4}. Although NeRF and its extensions can render high-quality images, their training time is excessively long, and the rendering speed is far from achieving real-time rendering ($\geq$ 30 fps). Recently, 3D Gaussian Splatting (3DGS)~\cite{3DGS} has made a significant impact in NVS due to its real-time rendering speed, high-quality rendering results and competitive training times. Unlike NeRF, 3DGS represents scenes with a set of Gaussian ellipsoids explicitly. By projecting each Gaussian ellipsoid onto the image plane and using $\alpha$-blending for rendering, the properties of each Gaussian ellipsoid, position, pose, scale, transparency, and color, are optimized based on a multi-view photometric loss.

Although 3D Gaussian Splatting has shown impressive results, due to the local affine approximation of the projection transformation~\cite{EWAsplatting} at the center of each Gaussian ellipsoid during rendering, errors are inevitably introduced, negatively affecting rendering quality and scene consistency. We observed blurriness and artifacts in distant objects in the scene, which we attribute to larger errors in the approximated projection transformation as the distance from the Gaussian ellipsoid center grows, particularly when distant Gaussian ellipsoids are generally large. Additionally, while the details rendered in the training set are of high quality, the results in the test set show a decline, likely due to the lack of scene consistency on account of the approximated projection transformation.

To solve this problem, we propose an ellipsoid-based projection method. Our core idea is to calculate the ellipse equation projected onto the image plane based on the work of David Eberly~\cite{project_ellipsoid}, given the equation of the ellipsoid and the image plane. We first derive the gaussian ellipsoid equation from the covariance matrix of the 3D Gaussian function. Then we find the equation of the cone formed by lines that pass through the camera's origin and are tangent to the ellipsoid surface. Finally, we determine the intersection line of this cone with the image plane, which gives us the projected ellipse equation.

The following experiments demonstrate that there are two types of Gaussian ellipsoids that can cause the training process to diverge and negatively impact the system. The first type has the camera's origin inside the ellipsoid, where no lines through the camera's origin can be tangent to it. The second type consists of Gaussian ellipsoids that have a portion below $z=0$ plane in the camera space, the projection of these ellipsoids results in hyperbola or parabola~\cite{project_ellipsoid} rather than an ellipse. To avoid negatively impacting the system, we designed filtering algorithms specifically for these two types of Gaussian ellipsoids. Extensive experiments demonstrated that our method not only improves rendering quality compared to 3DGS but also further accelerates rendering speed.

\noindent In summary, we make the following contributions:
\begin{itemize}
\item We proposed an \textbf{ellipsoid-based projection method} to eliminate the negative impact on rendering quality and scene consistency caused by approximating the projection transformation using the Jacobian of its affine approximation in 3DGS.
\item We design a \textbf{pre-filtering strategy} for Gaussian ellipsoids that cannot be projected before the rendering process, enhancing the system's robustness and contributing to faster rendering speed.
\item Experiments conducted on challenging benchmark datasets demonstrated that our method surpasses 3DGS in both rendering quality and speed.
\item Our ellipsoid-based projection method shows improvement results on 3DGS and its extensions, and can be easily applied to them, requiring only few changes to the original code.
\end{itemize}

%% file: sec/2.RelatedWorks.tex
\section{Related Work}

\subsection{Novel View Synthesis}
\label{sec:2.1}

The goal of Novel View Synthesis (NVS) is to generate images from new perspectives different from those of the captured images. There have been notable progressions in NVS, especially since the introduction of Neural Radiance Fields (NeRF)~\cite{NeRF}. NeRF uses a Multi-Layer Perceptron (MLP) to represent geometry and view-dependent appearance, optimized through volume rendering techniques \cite{volume_rendering1,volume_rendering2,volume_rendering3,volume_rendering4} to achieve high-quality rendering results. However, during the rendering process, the geometric and appearance information for each point along the ray must be obtained through a complex MLP network, resulting in slow rendering speed. Subsequent work has utilized distillation and baking techniques \cite{distillation_baking1,distillation_baking2,distillation_baking3,distillation_baking4,distillation_baking5} to increase NeRF's rendering speed, usually at a cost of lower rendering quality. Furthermore, the training speed and representational power of NeRF have been enhanced using
feature-grid based scene representations \cite{grid1,grid2,grid3,grid4,grid5}. In addition, some extended works have further improved rendering quality \cite{render_quality1,render_quality2,render_quality3} achieving state-of-the-art performance.

Recently, 3D Gaussian splatting~\cite{3DGS} has emerged as a method for representing intricate scenes using 3D Gaussian ellipsoids. This approach has shown remarkable results in NVS, enabling efficient optimization and rendering high quality images in real-time. This method has rapidly been extended to various domains \cite{domains1,domains2,domains3,domains4,domains5,domains6,domains7,domains8}. Most related work enhances the rendering quality of 3D Gaussian splatting through anti-aliasing \cite{anti-aliasing1,anti-aliasing2,anti-aliasing3}, combining it with other techniques like NeRF \cite{NeRF_GS1,NeRF_GS2,NeRF_GS3}, or proposing new Gaussian augmentation and reduction strategies. Additionally, others improve rendering quality by regularization \cite{anti-aliasing2}, or modifying Gaussian properties \cite{property1,property2}. Recently, some works \cite{ray1,ray2,ray3,ray4} has adopted ray tracing to replace the rasterization-based rendering method of 3DGS to improve rendering quality, but this has also led to a significant decrease in rendering speed. In this work, we propose a new projection method to optimize the rendering process and enhance rendering quality.

\subsection{Primitive-based Differentiable Rendering}
\label{sec:2.2}

Primitive-based rendering techniques, which project geometric primitives onto the image plane, have been widely studied for their efficiency \cite{Primitive1,Primitive2,Primitive3,Primitive4,Primitive5,Primitive6}. Differentiable point-based rendering methods provide significant flexibility in representing complex structures, making them ideal for novel view synthesis. Notably, NPBG~\cite{NPBG} rasterizes features from point clouds onto the image plane and then employs a convolutional neural network for RGB image prediction. DSS~\cite{DSS} aims to optimize oriented point clouds derived from multi-view images under specific lighting conditions. Pulsar~\cite{Pulsar} presents a tile-based acceleration structure to enhance the efficiency of rasterization. More recently, 3DGS~\cite{3DGS} projects anisotropic Gaussian ellipsoids onto the image plane and uses $\alpha$-blending techniques to render the ellipses on the plane based on depth ordering, to further optimize the Gaussian ellipsoids. During the projection process, 3DGS introduces errors by approximating the projection transformation using the Jacobian of its affine approximation~\cite{EWAsplatting}. To fundamentally eliminate the negative impact of these errors, we propose an ellipsoid-based projection method to replace the Jacobian of the affine approximation of the projection transformation in 3DGS.

\subsection{Perspective Projection of an Ellipsoid}
\label{sec:2.3}

In 1999, David Eberly et al. provided a comprehensive process for calculating the projection of an ellipsoid onto a given plane~\cite{project_ellipsoid}. This algorithm has primarily been used for the localization of 3D objects \cite{localization1,localization2,localization3} and for designing shadow models \cite{shadow1,shadow2}. In this work, based on David Eberly's findings, we propose a projection method for projecting 3D Gaussian ellipsoids onto the image plane. The representation of the ellipsoid is achieved by using the inverse of the covariance matrix of 3D gaussian function. After obtaining the ellipse equation, we further convert the result from the camera space to the image space. The projection algorithm~\cite{project_ellipsoid} proposed by David Eberly et al. has certain preconditions to ensure the projection of the ellipsoid onto the image plane is an ellipse. Therefore, we designed a pre-filtering strategy to eliminate Gaussian ellipsoids that do not meet these preconditions before the rendering process.

%% file: sec/3.Preliminaries.tex
\section{Preliminaries}

In this section, we first introduce the scene representation method of 3DGS~\cite{3DGS}, along with the rendering and optimization process in \cref{sec:3.1}. Subsequently, in \cref{sec:3.2}, we provide some algebraic details of the method proposed by David Eberly et al. in 1999~\cite{project_ellipsoid} for projecting ellipsoids onto the image plane.

\subsection{3D Gaussian Splatting}
\label{sec:3.1}

3D Gaussian splatting~\cite{3DGS} represents three-dimensional scenes using Gaussian ellipsoids as scene primitives. The geometric features of each Gaussian ellipsoid are represented by $\mathbf{p}\in \mathbb{R}^{3\times 1}$ and the covariance matrix $\mathbf{\Sigma}\in \mathbb{R}^{3\times 3}$. Where, $\mathbf{p}$ controls the geometric position of the Gaussian ellipsoid, and $\mathbf{\Sigma}$ controls its shape. The covariance matrix $\mathbf{\Sigma}$ consist of a rotation matrix $\mathbf{R}$ and a scaling matrix $\mathbf{S}$, which controls the orientation and the scale of the Gaussian ellipsoid, respectively. 

\begin{equation}
  \mathbf{\Sigma} = \mathbf{R}\mathbf{S}\mathbf{S}^{T}\mathbf{R}^{T}
  \label{eq:covariance matrix}
\end{equation}

The appearance components of each Gaussian ellipsoid are represented by $\alpha \in [0,1]$ and spherical harmonics (SH). $\alpha$ controls the transparency of the Gaussian ellipsoid, and SH controls the color distribution of the ellipsoid from different viewpoints.

3DGS generates a set of sparse point clouds through Structure from Motion (SfM) to initialize the Gaussian ellipsoids, with centers of the ellipsoids corresponding to the point clouds. For further rendering, the covariance matrix $\mathbf{\Sigma}$ of a Gaussian ellipsoid is transformed into the camera space using a viewing transformation $\mathbf{W}$. Then, the covariance matrix is projected onto the image plane via the projection transformation, which is approximated by the corresponding local affine approximation Jacobian matrix $\mathbf{J}$.

\begin{equation}
  \mathbf{\Sigma}^{\prime} = \mathbf{JW}\mathbf{\Sigma} \mathbf{W}^{T}\mathbf{J}^{T}
  \label{eq:transformation}
\end{equation}

\noindent After removing the third row and third column of $\mathbf{\Sigma}^{\prime}$, the 2D covariance matrix $\mathbf{\Sigma}^{2D}$ was obtained. By projecting the center of the Gaussian ellipsoid onto the image plane using the viewing transformation $\mathbf{W}$ and the projection transformation $\mathbf{P}$, the center of the ellipse $\mathbf{p}^{2D}$ can be obtained. 


The color $\mathbf{c}\in \mathbb{R}^{3\times 1}$ of a Gaussian ellipsoid at given viewpoint is calculated based on the spherical harmonic and the relative position of the ellipsoid to the camera's origin. Finally, $\alpha$-blending is used to render each pixel according to the sorted K Gaussian ellipsoids from near to far.

\begin{equation}
  \mathbf{C}(\mathbf{x}) = \sum_{k=1}^{K}\mathbf{c}_k\alpha_k\mathcal{G}_{k}^{2D}(\mathbf{x})\prod_{j=1}^{k-1}(1-\alpha_j\mathcal{G}_j^{2D}(\mathbf{x}))
  \label{eq:alpha blending}
\end{equation}

\noindent $\mathcal{G}^{2D}(\mathbf{x})$ is the probability density function of a 2D Gaussian distribution, which reflects the distance between the pixel $\mathbf{x}$ and the 2D Gaussian ellipse center $\mathbf{p}^{2D}$.

\begin{equation}
  \mathcal{G}_{k}^{2D}(\mathbf{x}) = e^{-\frac{1}{2}(\mathbf{x}-\mathbf{p}_k^{2D})^T(\mathbf{\Sigma}_k^{2D}+s\mathbf{I})^{-1}(\mathbf{x}-\mathbf{p}_k^{2D})}
  \label{eq:alpha blending}
\end{equation}

\noindent Adding $s\mathbf{I}$ to $\mathbf{\Sigma}^{2D}$ serves to prevent the projected 2D Gaussian ellipse from being smaller than one pixel.

Since the rendering process is fully differentiable, after calculating the photometric loss between each rendered image and the reference image, the parameters of each Gaussian ellipsoid can be optimized using gradient descent, thereby enhancing rendering quality. During the optimization process, based on the gradient of the 2D Gaussian ellipse's center $\mathbf{p}^{2D}$ in the image plane and the scaling matrix $\mathbf{S}$ of each Gaussian ellipsoid, 3D Gaussian splatting can adaptively adjust the number of Gaussian ellipsoids through clone, split, and prune strategies.

\subsection{Projection Algorithm for Ellipsoid}
\label{sec:3.2}

The standard equation of an ellipsoid is $(\mathbf{x}-\mathbf{c})^T\mathbf{A}(\mathbf{x}-\mathbf{c})=1$, where $\mathbf{c}$ is the center of the ellipsoid and $\mathbf{A}$ is a positive definite matrix. Given the camera origin $\mathbf{e}$ and the image plane $\mathbf{n}\cdot \mathbf{x}=k$, where $\mathbf{n}$ is a unit-length vector, $k$ is a constant. The task is to calculate the projection of the ellipsoid onto the image plane. 

The first step is to compute the cone that tightly bounds the ellipsoid. Consider
a ray $\mathbf{l}(t) = \mathbf{e} + t\mathbf{d}$, where $t > 0$ and $\mathbf{d}$ is a unit-length vector. The intersection of the ray and ellipsoid can be defined as

\begin{equation}
  (\mathbf{l}(t)-\mathbf{c})^T\mathbf{A}(\mathbf{l}(t)-\mathbf{c})=1,
  \label{eq:ray-ellipsoid intersection}
\end{equation}

\noindent or equivalently by

\begin{equation}
  \mathbf{d}^T\mathbf{A}\mathbf{d}t^2+2\mathbf{\delta}^T\mathbf{A}\mathbf{d}t+\mathbf{\delta}^T\mathbf{A}\mathbf{\delta}-1=0,
  \label{eq:ray-ellipsoid intersection-1}
\end{equation}

\noindent where $\mathbf{\delta}=\mathbf{e}-\mathbf{c}$. This is a quadratic equation in $t$. When the equation has a single real-valued root, the ray is tangent to the ellipsoid. The condition for the quadratic equation to have a single real-valued root is

\begin{equation}
  (\mathbf{\delta}^T\mathbf{A}\mathbf{d})^2-(\mathbf{d}^T\mathbf{A}\mathbf{d})(\mathbf{\delta}^T\mathbf{A}\mathbf{\delta}-1)=0.
  \label{eq:single real-valued root}
\end{equation}

\noindent It can further be transformed into

\begin{equation}
  \mathbf{d}^T(\mathbf{A}^T\mathbf{\delta}\mathbf{\delta}^T\mathbf{A}-(\mathbf{\delta}^T\mathbf{A}\mathbf{\delta}-1)\mathbf{A})\mathbf{d}=0.
  \label{eq:single real-valued root1}
\end{equation}

\noindent The direction vector $\mathbf{d}$ of the ray can also be expressed as $\mathbf{d}=\frac{\mathbf{x}-\mathbf{e}}{|\mathbf{x}-\mathbf{e}|}$. Substituting this into the equation yields

\begin{equation}
  (\mathbf{x}-\mathbf{e})^T(\mathbf{A}^T\mathbf{\delta}\mathbf{\delta}^T\mathbf{A}-(\mathbf{\delta}^T\mathbf{A}\mathbf{\delta}-1)\mathbf{A})(\mathbf{x}-\mathbf{e})=0.
  \label{eq:cone equation}
\end{equation}

\noindent This equation describes a cone with vertex at $\mathbf{e}$. 

Next, by solving the combined equations of the cone and the image plane, the projection of the ellipsoid onto the image plane can be obtained.

%% file: sec/4.ProposedMethod.tex
\section{Proposed Method}

Due to errors introduced by approximating the projection transformation using the Jacobian of its affine approximation in 3D Gaussian splatting~\cite{3DGS}, we propose an ellipsoid-based projection method to eliminate the errors and enhance rendering quality and scene consistency. In \cref{sec:4.1}, we explain why errors are inevitably introduced in the projection process of 3DGS, followed by the presentation of our ellipsoid-based projection method and its theoretical derivation. Next, in \cref{sec:4.2}, we identify two types of Gaussian ellipsoids that cannot be rendered with the proposed projection method and present a pre-filtering strategy to filter out these two types of Gaussian ellipsoids.

\subsection{Ellipsoid-based Projection Method}
\label{sec:4.1}

Gaussian functions are closed under linear transformations but not under nonlinear transformations, which means the result of nonlinear transformations no longer represent an ellipsoid (or ellipse). Since projection transformation is nonlinear, 3DGS~\cite{3DGS} adopts a local affine approximation of the projection transformation $\mathbf{P}$ at the center of the Gaussian ellipsoid to obtain the Jacobian matrix $\mathbf{J}$. Using the Jacobian of the affine approximation of the projection transformation will inevitably introduce errors at positions other than the center of the Gaussian ellipsoid during the projection process, and farther distance from the center gives larger error.

\begin{figure}[t]
  \centering
   \includegraphics[width=0.9\linewidth]{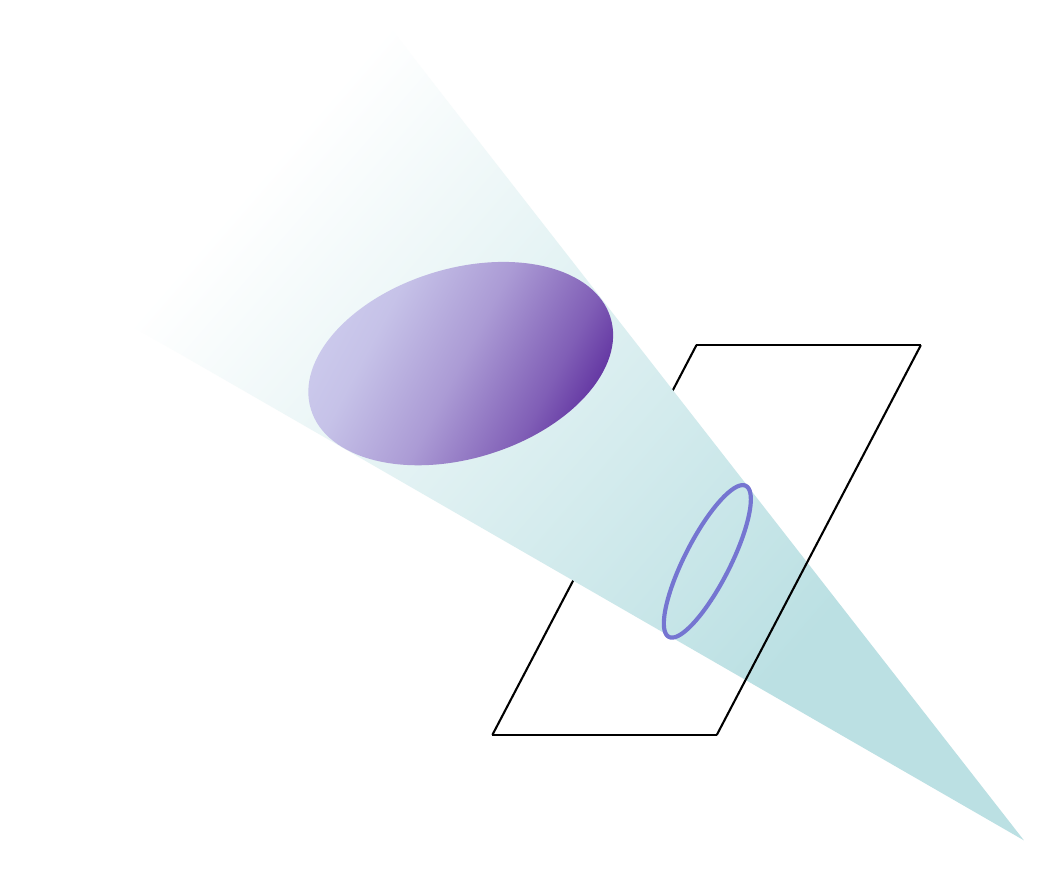}

   \caption{\textbf{Ellipsoid-based Projection Method}. We first derive the ellipsoid equation based on the covariance matrix of the 3D Gaussian function. Then, using the method in \cref{sec:3.2}, we obtain the equation of a cone with its vertex at the camera origin and tangent to the ellipsoid. Finally, we calculate the intersection line between the cone and the image plane, which gives the equation of the projection of the ellipsoid.}
   \label{fig:projection equation}
\end{figure}

Based on the method introduced by David Eberly et al.~\cite{project_ellipsoid}, we designed an ellipsoid-based projection method for 3DGS. To simplify the computation, we will project the Gaussian ellipsoid in the camera space $\mathcal{C}\{x,y,z\}$. First, we need to derive the equation of the Gaussian ellipsoid based on 3D covariance matrix $\mathbf{\Sigma}$. In 3D Gaussian Splatting~\cite{3DGS}, the author obtains the corresponding Gaussian ellipse equation by setting the exponential part of the 2D Gaussian function equal to $3^2$ (satisfying the $3\sigma$ principle). Similarly, for the 3D Gaussian function, we also set its exponential part equal to $3^2$:

\begin{equation}
  (\mathbf{x}-\mathbf{p}_c)^T\mathbf{\Sigma}_c^{-1}(\mathbf{x}-\mathbf{p}_c) = 9,
  \label{eq:ellipsoid equation 3dgs}
\end{equation}

\noindent where $\mathbf{p}_c$ is the center of Gaussian ellipsoid in camera space and $\mathbf{\Sigma}_c$ is the 3D covariance matrix $\mathbf{\Sigma}$ in camera space.

In camera space, the camera origin $\mathbf{c}=[0,0,0]$. According to \cref{eq:ray-ellipsoid intersection}, \cref{eq:single real-valued root} and \cref{eq:cone equation}, the equation of the cone can be obtained.

\begin{equation}
  \mathbf{x}^T(\mathbf{\Sigma}_c^{-T}\mathbf{p}_c\mathbf{p}_c^T\mathbf{\Sigma}_c^{-1}-(\mathbf{p}_c^T\mathbf{\Sigma}_c^{-1}\mathbf{p}_c-9)\mathbf{\Sigma}_c^{-1})\mathbf{x}=0
  \label{eq:cone equation 3dgs}
\end{equation}

\noindent Setting the third element of $\mathbf{x}=[x,y,z]$ to 1 gives the intersection between the cone and the $z=1$ plane. Thus, the equation of this intersection is an ellipse equation that depends only on $x$ and $y$. 

For one point $\mathbf{x}^{2d}=[x,y]$ in the $z=1$ plane, scaling and translation are required to convert it into a point $\mathbf{x}^{img}=[x^{img},y^{img}]$ in the image plane $\mathcal{I}\{x^{img},y^{img}\}$. The corresponding relationship is

\begin{equation}
  \begin{cases}
  x^{img}=f_xx+\frac{1}{2}w \\[8pt]
  y^{img}=f_yy+\frac{1}{2}h,
  \end{cases}
  \label{eq:img plane}
\end{equation}

\noindent where $f_x$ and $f_y$ are the camera intrinsic parameters, and $w$ and $h$ represent the width and height of the image, respectively. 

We substitute $x^{img}$ and $y^{img}$ into the ellipse equation to obtain the equation of the ellipse on the image plane

\begin{equation}
  (\mathbf{x}^{img}-\mathbf{p}_{img})^T\mathbf{\Sigma}_{img}^{-1}(\mathbf{x}^{img}-\mathbf{p}_{img}) = 9,
  \label{eq:final ellipse equation}
\end{equation}

\noindent where $\mathbf{\Sigma}_{img}^{-1}\in\mathbb{R}^{2\times 2}$ is the inverse of the second-order covariance matrix corresponding to the ellipse and $\mathbf{p}_{img}\in\mathbb{R}^{2\times 1}$ is the center of the ellipse.

We find that the center of the ellipse calculated from our projection transformation has a slight difference compared to the result obtained by directly projecting the center of the Gaussian ellipsoid onto the image plane. This indicates that the center of the ellipse is no longer the projection of the center of the Gaussian ellipsoid, which also explains why the projected result remains an ellipse, but it is not possible to find an affine transformation that directly converts the 3D Gaussian covariance matrix into a 2D one.

\subsection{Pre-filtering Strategy}
\label{sec:4.2}

\begin{figure}[t]
  \centering
   \includegraphics[width=0.9\linewidth]{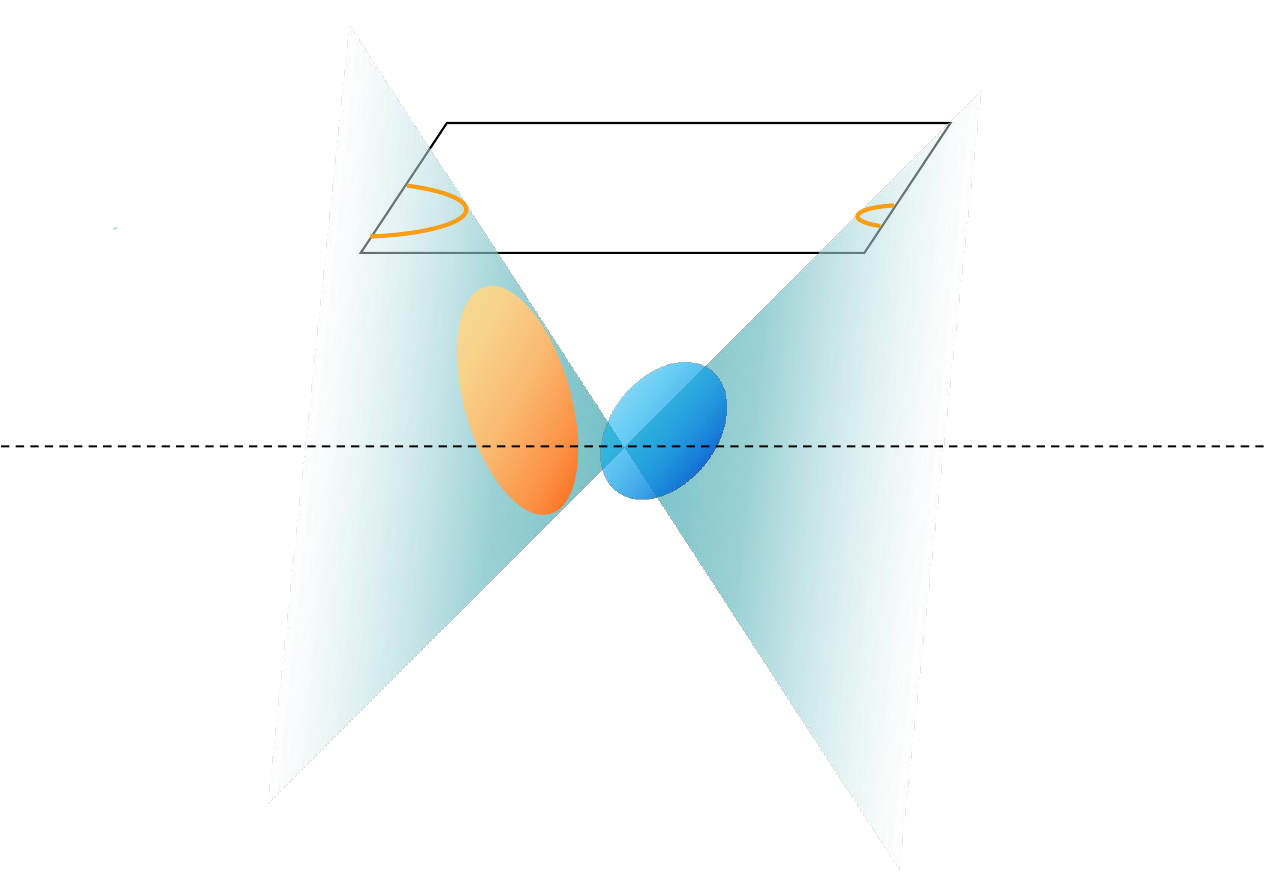}

   \caption{\textbf{Pre-filtering Strategy}. There are two types of Gaussian ellipsoids that need to be filtered out in advance. Otherwise, the system may fail to converge. The first type is Gaussian ellipsoids that contain the camera origin within them, represented by the blue ellipsoid in the figure. The second type consists of Gaussian ellipsoids with portions located below the $z=0$ plane in the camera space. The projection of these ellipsoids results in parabolas or hyperbolas, as shown by the orange ellipsoid in the figure.}
   \label{fig:bad gaussians}
\end{figure}

For the ellipsoid-based projection method we proposed, there are two types of Gaussian ellipsoids that cannot be rendered~\cite{project_ellipsoid} and need to be filtered out beforehand, otherwise, they will cause the training process to diverge and negatively impact the system. The first type is the Gaussian ellipsoid with the camera origin inside it, for which there is no line passing through the camera origin that is tangent to it, as the blue Gaussian ellipsoid in \cref{fig:bad gaussians}. The second type is the Gaussian ellipsoid that has a portion below $z=0$ plane in the camera space. Using our projection method, the projection of this type of Gaussian ellipsoid onto the image plane results in a hyperbola, not an ellipse, as the orange Gaussian ellipsoid in \cref{fig:bad gaussians}.

By substituting the camera origin $\mathbf{c}=[0,0,0]$ into the left side of \cref{eq:ellipsoid equation 3dgs}, if the result is less than or equal to 9 

\begin{equation}
  (\mathbf{c}-\mathbf{p}_c)^T\mathbf{\Sigma}_c^{-1}(\mathbf{c}-\mathbf{p}_c) \le 9,
  \label{eq:filter1}
\end{equation}

\noindent it indicates that the camera origin lies inside or on the surface of the ellipsoid. Such ellipsoids belong to the first type of Gaussian ellipsoids.

The equation of the ellipsoid is $\mathcal{E}(\mathbf{x})=0$, where $\mathcal{E}(\mathbf{x})=(\mathbf{x}-\mathbf{p}_c)^T\mathbf{\Sigma}_c^{-1}(\mathbf{x}-\mathbf{p}_c)-9$. For the second type of Gaussian ellipsoid, we first need to find the lowest point of the ellipsoid surface in the direction of the z-axis, where the gradient direction of the ellipsoid surface is perpendicular to the plane $z=0$, satisfying $\nabla \mathcal{E}(\mathbf{x})=k\mathbf{n}$, where $\mathbf{n}=[0,0,1]$. By combining the ellipsoid equation and the gradient equation

\begin{equation}
  \begin{cases}
  \mathcal{E}(\mathbf{x})=0 \\[8pt]
  \frac{\partial\mathcal{E}(\mathbf{x})}{\partial x}=0 \\[8pt]
  \frac{\partial\mathcal{E}(\mathbf{x})}{\partial y}=0,
  \end{cases}
  \label{eq:img plane}
\end{equation}

\noindent we can determine the coordinate $[x,y,z]$ of that point. If $z\le 0$, it needs to be filtered. When $z=0$, the projection result is a parabola, while when $z< 0$, the projection result is a hyperbola.

%% file: sec/5.Experiments.tex
\section{Experiments}

\input{tables/ourVS3dgs}

\input{tables/ourVS3dgs_tnt}

We first introduce datasets used in our experiments and implementation details in \cref{sec:5.1}. In \cref{sec:5.3}, we evaluate our method on three datasets and compare it with 3DGS~\cite{3DGS} and other state-of-the-art. Subsequently, in \cref{sec:5.4}, we applied the ellipsoid-based projection method in Mip-Splatting~\cite{anti-aliasing1} and compared it with the original method. Finally, in \cref{sec:5.5} we analyze the limitations of our method and explore ways for future improvement.

\subsection{Datasets and Implementation}
\label{sec:5.1}

\textbf{Datasets} \quad For training and testing, we perform experiments on images from a total of 16 real-world scenes. Specifically, we evaluate our ellipsoid-based projection method on 7 scenes from Mip-NeRF360 dataset~\cite{render_quality2}, 7 scenes from Tanks\&Temples dataset~\cite{Tanks&temples}, and 2 scenes from Deep Blending dataset~\cite{DeepBlending}. The selected scenes showcase diverse styles, including bounded indoor environments and unbounded outdoor ones. To divide the datasets into training and testing sets, we follow the method used in 3DGS~\cite{3DGS}, assigning every 8th photo to the test set. For images in Mip-NeRF360 and Tanks\&Temples datasets, we use $\frac{1}{2}$ original resolution for training and rendering.

\noindent \textbf{Implementation} \quad Our method is built on the open-source code of 3DGS~\cite{3DGS}. Following the 3DGS framework, we set the number of training iterations to 30k for all scenes, using the same loss function and densification strategy as 3DGS, with all hyperparameters remaining consistent. We only modified the CUDA kernels in the projection parts of the forward and backward processes, replacing the Jacobian of the affine approximation of the projection transformation in 3DGS with our ellipsoid-based projection method. In the comparison experiments with Mip-Splatting~\cite{anti-aliasing1}, we integrated the ellipsoid-based projection method into Mip-Splatting in the same way. All our experiments are conducted on a single GTX RTX3090 GPU.

\begin{figure*}[htbp]
    \centering
    \includegraphics[width=1.\textwidth]{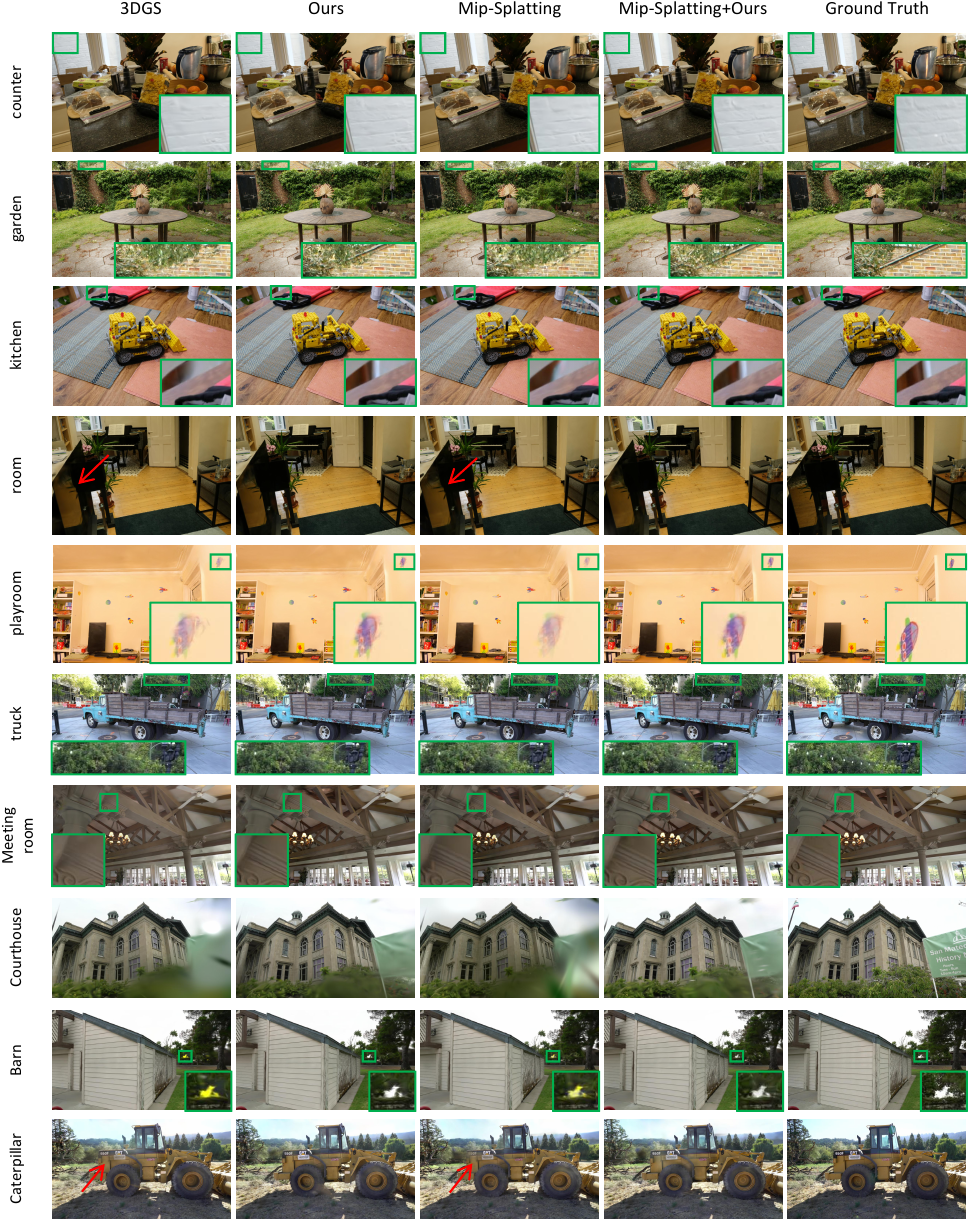}
    \caption{We demonstrated the rendering results of applying our ellipsoid-based projection method to 3DGS and Mip-Splatting, resulting in less blur and artifact and better scene consistency.}
    \label{fig:Qualitative Results}
\end{figure*}

\input{tables/ourVSmip}

\subsection{Comparisons with 3DGS}
\label{sec:5.3}

Similar to 3DGS~\cite{3DGS}, we use PSNR, SSIM, and LPIPS as metrics to evaluate rendering quality. Additionally, we assess the rendering speed of our method using Frames Per Second (FPS), which we calculate by averaging the rendering time for all images in each scene to obtain the FPS. The results of the rendering quality evaluation are shown in \cref{tab:ourVS3dgs} and \cref{tab:ourVS3dgs_tnt}, with the metrics for each dataset calculated by averaging across all scenes in it. For all scenes in Mip-NeRF360 dataset~\cite{render_quality2}, our method outperforms 3DGS in PSNR, SSIM, and LPIPS. In some scenes within the Tanks\&Temples dataset~\cite{Tanks&temples}, our results show a slight gap in PSNR compared to 3DGS. In the two scenes from the Deep Blending dataset~\cite{DeepBlending}, our method scores slightly lower than 3DGS in SSIM. Across all 16 scenes, however, our method achieves higher LPIPS scores than 3DGS, indicating that our rendered results align more closely with human perception. \cref{fig:Qualitative Results} shows a visual comparison between our method and 3DGS for several test views. Compared to 3DGS, our results have reduced blur (as shown in \textit{counter, kitchen, room} in \cref{fig:Qualitative Results}) and artifacts (as shown in \textit{Barn, Caterpillar} in \cref{fig:Qualitative Results}), with better scene consistency (as shown in \textit{garden, Courthouse} in \cref{fig:Qualitative Results}) and details (as shown in \textit{palyroom, Meetingroom, truck} in \cref{fig:Qualitative Results}). Furthermore, our method significantly outperforms 3DGS in rendering speed, likely because we pre-filter Gaussian ellipsoids before rendering. In addition to the Gaussian ellipsoids mentioned in \cref{sec:4.2} that prevent the training process from converging, we also filter out Gaussian ellipsoids outside the cone of vision that would not be rendered. Additionally, our method significantly outperforms Plenoxels~\cite{Plenoxels} and Instant-NGP~\cite{grid1} in rendering quality. Except for PSNR, which is lower than the best-performing Mip-NeRF360~\cite{render_quality2} on their dataset, our method surpasses Mip-NeRF360 across all other metrics.

\subsection{Comparisons with Mip-Splatting}
\label{sec:5.4}

To further demonstrate the effectiveness of our algorithm, we applied our ellipsoid-based projection method to Mip-Splatting~\cite{anti-aliasing1} and compared it with the original Mip-Splatting method using PSNR, SSIM, and LPIPS metrics. We utilized Mip-Splatting's core methods, 3D-filter and 2D-filter, but retained the Gaussian densification strategy consistent with 3DGS rather than adopting the newly proposed strategy from Mip-Splatting. In our experiments, we observed that the Gaussian densification strategy in Mip-Splatting led to a significant increase in the number of Gaussians for the original method, but had no noticeable effect when used with our projection transformation. Therefore, we conclude that this strategy is specific to Mip-Splatting. We set the 2D-filter parameter to 0.25, making the kernel size a $3\times3$ pixel region on the image. As shown in the results in \cref{tab:ourVSmip}, applying the ellipsoid-based projection method to Mip-Splatting improved rendering quality. Similar to the results in \cref{sec:5.3}, our method achieved overall metric improvements on the Mip-NeRF360 dataset, showed a slight decrease in PSNR on the Tanks\&Temples dataset, and a slight decrease in SSIM on the Deep Blending dataset. Visual results in \cref{fig:Qualitative Results} show that, similar to 3DGS~\cite{3DGS} in \cref{sec:5.3}, Mip-Splatting with the ellipsoid-based projection method also reduces blur and artifacts, while enhancing scene consistency and details.

\subsection{Limitations}
\label{sec:5.5}

Our method does not surpass 3DGS~\cite{3DGS} in every metric across all scenes, indicating there is still room for further improvement. In \cref{sec:5.4}, we observed that Mip-Splatting's~\cite{anti-aliasing1} Gaussian densification strategy had limited impact on our method, suggesting that we may not be using the most suitable Gaussian densification strategy. If a more appropriate strategy could be identified, rendering quality might be further enhanced. In the experiments in \cref{sec:5.3} and \cref{sec:5.4}, we used fixed values for the filter parameters. However, we found that our method is more sensitive to filter parameters than 3DGS, whether for the screen-space dilation filter in 3DGS or the 2D mip filter in Mip-Splatting. Additionally, the optimal filter parameters vary across different scenes, making parameter selection crucial. Since filter parameters are continuous values, it is difficult to identify optimal values through repeated experiments. One potential solution is to include the filter parameters as optimizable variables and optimize them during training.

%% file: tables/ourVS3dgs.tex
\begin{table*}[h]
\centering
\resizebox{\textwidth}{!}{
\begin{tabular}{l|c c c c c|c c c c c|c c c c c}
{Dataset} & \multicolumn{5}{c|}{Mip-NeRF360 (7 scenes)} & \multicolumn{5}{c|}{Tanks\&Temples (2 scenes)} & \multicolumn{5}{c}{Deep Blending (2 scenes)} \\

{Method$\vert$Metric} & {PSNR$\uparrow$} & {SSIM$\uparrow$} & {LPIPS$\downarrow$} & {FPS} & {Mem} & {PSNR$\uparrow$} & {SSIM$\uparrow$} & {LPIPS$\downarrow$} & {FPS} & {Mem} & {PSNR$\uparrow$} & {SSIM$\uparrow$} & {LPIPS$\downarrow$} & {FPS} & {Mem} \\
\hline
Plenoxels & 23.62 & 0.670 & 0.443 & 6.79 & 2.1GB & 21.08 & 0.719 & 0.379 & 13.0 & 2.3GB & 23.06 & 0.795 & 0.510 & 11.2 & 2.7GB \\
INGP-Base & 26.43 & 0.725 & 0.339 & 11.7 & 13MB & 21.72 & 0.723 & 0.330 & 17.1 & 13MB & 23.62 & 0.797 & 0.423 & 3.26 & 13MB \\
INGP-Big & 26.75& 0.751 & 0.302 & 9.43 & 48MB & 21.92 & 0.745 & 0.305 & 14.4 & 48MB & 24.96 & 0.817 & 0.390 & 2.79 & 48MB \\
M-NeRF360 & \cellcolor{tablered}29.09 & \cellcolor{yellow}0.842 & \cellcolor{yellow}0.210 & 0.06 & 8.6MB & \cellcolor{yellow}22.22 & \cellcolor{yellow}0.759 & \cellcolor{yellow}0.257 & 0.14 & 8.6MB & \cellcolor{yellow}29.40 & \cellcolor{yellow}0.901 & \cellcolor{yellow}0.245 & 0.09 & 8.6MB \\
3DGS & \cellcolor{yellow}28.78 & \cellcolor{orange}0.857 & \cellcolor{orange}0.210 & 114 & 711MB & \cellcolor{tablered}23.64 & \cellcolor{orange}0.848 & \cellcolor{orange}0.177 & 168 & 442MB & \cellcolor{orange}29.60 & \cellcolor{orange}0.904 & \cellcolor{orange}0.244 & 130 & 682MB \\

\hline

Ours  & \cellcolor{orange}28.82 & \cellcolor{tablered}0.858 & \cellcolor{tablered}0.208 & 120 & 733MB & \cellcolor{orange}23.57 & \cellcolor{tablered}0.849 & \cellcolor{tablered}0.176 & 269 & 448MB   & \cellcolor{tablered}29.66 & \cellcolor{tablered}0.904 & \cellcolor{tablered}0.242 & 143 & 706MB\\
\end{tabular}}

\caption{We evaluated our method on the Mip-NeRF360, Tank\&Temple, and Deep Blending datasets by comparing it with previous approaches. The results for Plenoxels~\cite{Plenoxels}, InstantNGP~\cite{grid1}, and Mip-NeRF360~\cite{render_quality2} were obtained directly from the 3DGS~\cite{3DGS}. Note that the results for all Mip-NeRF360 datasets in the table are calculated based on 7 scenes, excluding scene \textit{flowers} and \textit{treehill}. For a fair comparison with previous methods, the results for Tanks\&Temples were calculated based on scene \textit{truck} and \textit{train}, and the full results for all 7 scenes are shown in \cref{tab:ourVS3dgs_tnt}.}
\label{tab:ourVS3dgs}
\end{table*}

%% file: tables/ourVS3dgs_tnt.tex
\begin{table}[h]
\centering
\resizebox{0.5\textwidth}{!}{
\begin{tabular}{l|c c c c c}
{Dataset} & \multicolumn{5}{c}{Tanks\&Temples (7 scenes)}\\

{Method$\vert$Metric} & {PSNR$\uparrow$} & {SSIM$\uparrow$} & {LPIPS$\downarrow$} & {FPS} & {Mem}  \\
\hline

3DGS & 24.423 & 0.845 & 0.183 & 200 & 366MB \\
Ours & \textbf{24.424} & \textbf{0.846} & \textbf{0.181} & 351 & 374MB\\

\end{tabular}}

\caption{Comparison between our method and the 3DGS across all 7 scenes of the Tank\&Temple dataset.}
\label{tab:ourVS3dgs_tnt}

\end{table}

%% file: tables/ourVSmip.tex
\begin{table*}[h]
\centering
\resizebox{\textwidth}{!}{
\begin{tabular}{l|c c c|c c c|c c c}
{Dataset} & \multicolumn{3}{c|}{Mip-NeRF360 (7 scenes)} & \multicolumn{3}{c|}{Tanks\&Temples (7 scenes)} & \multicolumn{3}{c}{Deep Blending (2 scenes)} \\

{Method$\vert$Metric} & {PSNR$\uparrow$} & {SSIM$\uparrow$} & {LPIPS$\downarrow$} & {PSNR$\uparrow$} & {SSIM$\uparrow$} & {LPIPS$\downarrow$} & {PSNR$\uparrow$} & {SSIM$\uparrow$} & {LPIPS$\downarrow$}\\
\hline
Mip-Splatting & 28.87 & 0.858 & 0.210 & \textbf{24.60} & 0.847 & 0.185 & 29.87 & \textbf{0.911} & 0.192\\
Mip-Splatting+Ours & \textbf{28.94} & \textbf{0.859} & \textbf{0.208} & 24.55 & \textbf{0.847} & \textbf{0.183} & \textbf{29.90} & 0.911 & \textbf{0.191}\\
\end{tabular}}

\caption{Evaluate the effectiveness of our method on mip-splatting across all 16 scenes in the three datasets.}
\label{tab:ourVSmip}
\end{table*}

%% file: sec/6.Conclusion.tex
\section{Conclusion}

We propose an ellipsoid-based projection method that avoids errors introduced by the affine approximation of the projection transformation used in 3DGS. Additionally, we introduce a pre-filtering strategy to remove Gaussian ellipsoids that negatively impact the system or are outside the cone of vision, enhancing system robustness. Comparative experiments with 3DGS demonstrate that our method significantly improves rendering quality in both complex indoor and outdoor scenes, and accelerates rendering speed at the same time. By integrating our method with Mip-Splatting, rendering quality was further improved, proving the versatility of our method and its ease of adaptation to any work based on 3DGS.